# Interactive Image-Based Aphid Counting in Yellow Water Traps under Stirring Actions

Xumin Gao[1*], Mark Stevens[2], Grzegorz Cielniak[1]

[1] Lincoln Centre for Autonomous Systems, University of Lincoln, Lincoln, UK

[2] British Beet Research Organisation, Colney Lane, Norwich, UK

25766099@students.lincoln.ac.uk

**Abstract.** The current vision-based aphid counting methods in water traps suffer from under-counts caused by occlusions and low visibility arising from dense aggregation of insects and other objects. To address this problem, we propose a novel aphid counting method through interactive stirring actions. We use interactive stirring to alter the distribution of aphids in the yellow water trap and capture a sequence of images which are then used for aphid detection and counting through an optimized small object detection network based on Yolov5. We also propose a counting confidence evaluation system to evaluate the confidence of counting results. The final counting result is a weighted sum of the counting results from all sequence images based on the counting confidence. Experimental results show that our proposed aphid detection network significantly outperforms the original Yolov5, with improvements of 33.9% in AP@0.5 and 26.9% in AP@[0.5:0.95] on the aphid test set. In addition, the aphid counting test results using our proposed counting confidence evaluation system show significant improvements over the static counting method, closely aligning with manual counting results.

## 1    Introduction

Aphids can damage crops by feeding and transmitting viruses, resulting in economic losses. In Australia, aphid-related crop losses are estimated at $241 million from feeding and $482 million from virus transmission annually [1]. Therefore, timely monitoring and control of aphid populations are crucial. However, commonly used manual aphid counting is labor-intensive and time-consuming. Some research is currently focused on automatic aphid counting using image recognition. However, these works [3-7] often rely on static images, which can be inaccurate due to occlusion from aggregation or foreign objects, leading to missed counts. To this end, this paper proposes an automatic aphid counting method through interactive stirring actions. First, stirring alters the distribution of aphids in the yellow water trap, making some of the occluded individuals visible for detection. Second, given the potential for false and missed detections of detection network, we propose a counting confidence evaluation system to evaluate the confidence of multiple counting results derived from detection results on sequential images captured during stirring. And the final aphid count is calculated as a weighted sum of the multiple counting results based on counting confidence. Rather than relying on the maximum count value from these counting results as the final aphid count, this approach allows for a more accurate estimation. For aphid detection, we propose a small object detection network based on Yolov5 [2]. As this work is still in progress, we will focus on demonstrating its design principles.

## 2    Related work

Current research on automatic aphid counting encompasses image processing, traditional machine learning, and deep learning methods. Shen et al. [3] convert images to HSI color space to separate leaves from aphids and use morphological operations as post-processing for aphid counting. Suo et al. [4] employed image segmentation and contour extraction to count aphids on yellow pasteboards. The common drawback of both [3] and [4] is their assumption that all pests on the leaves or pasteboards are aphids, as they can't distinguish between aphids and other pests. Lins et al. [5] also used image segmentation and contour extraction to count aphids, but their method requires an ideal setup with washed aphids in a transparent petri dish. Liu et al. [6] used HOG features with an SVM classifier for aphid recognition, achieving a test accuracy of 86.81%. However, this method relies heavily on prior knowledge for feature design, making it less adaptable to changing environments. Júnior et al. [7] developed an automatic pest counting system using Mask R-CNN [8] to count aphids and other pests, achieving a determination coefficient of 0.81. But it also requires washing aphids and capturing images in a lab setting. Overall, it's important to note that all current automatic aphid counting methods rely on static counting, overlooking aphids that are occluded or in hidden areas, which leads to inaccurate counts.

Interactive perception is defined as enhancing perception by acquiring new sensory information through interaction with the environment. Cai et al. [9] investigated the link between object properties and grasping actions, achieving a 9.5% improvement in object property classification accuracy through interactive grasping. Le et al. [10] developed an active interactive perception method that improves object detection by altering the object's viewpoint. In tests with 35 objects, it correctly identified 34, compared to 23 with static recognition, greatly reducing misidentifications. Overall, interacting with objects improves recognition, especially for occluded ones.

## 3 Method

### 3.1 Overview of our proposed method

In this paper, we propose an automatic aphid counting method through interactive stirring actions, as shown in Fig. 1.

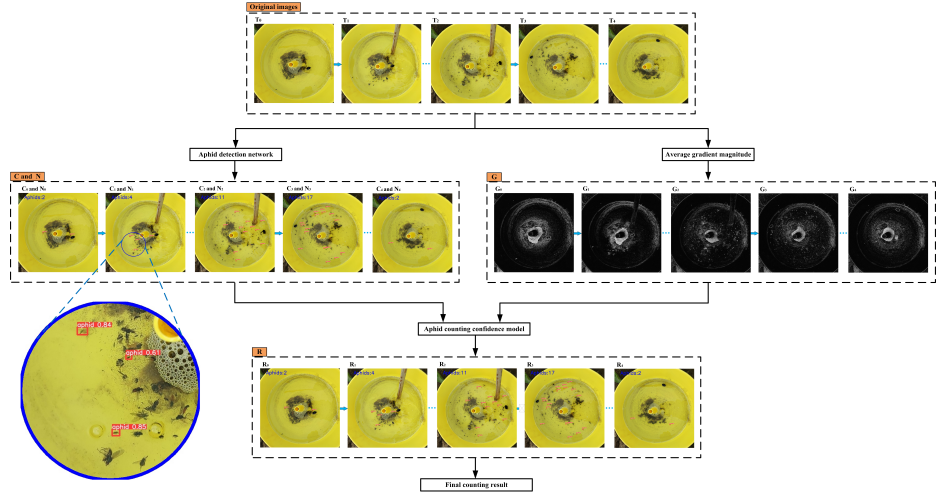

**Fig. 1.** Overview of our proposed method.

As illustrated in Fig. 1, an image is captured at $T_0$ before stirring begins. At $T_1$, the stirring tool is added to the yellow water trap, and stirring continues until $T_2$, when it stops. At $T_3$, the stirring tool is removed, and image capture continues until $T_4$, when the water surface is nearly calm, ultimately obtaining a complete sequence of images $Img = \{Img_1, Img_2, \dots Img_n\}$. Then, we input the sequence of images into the proposed aphid detection network to obtain initial counts. However, due to the detection model's inherent limitations, which may result in false or missed detections, leading to overestimation or underestimation. We conduct a confidence evaluation on each counting result from the sequence of images, with the metrics used for counting confidence evaluation including the detection confidence of aphid bounding boxes $C = \{C_1, C_2, \dots C_n\}$, the predicted number of aphids $N = \{N_1, N_2, \dots N_n\}$, and the average gradient magnitude $G = \{G_1, G_2, \dots G_n\}$, which represents the clarity of the images. Generally, as an image blurs, the average gradient magnitude decreases due to reduced brightness variations and loss of edges. Here, $C_i$ is the average confidence score of all aphid bounding boxes of the image $Img_i$, calculated as shown in Eq. (1). In Eq. (1), $N_i$ is the total number of aphid bounding boxes in the image $Img_i$, and $CS_j$ is the confidence score of the $j$-th detection box in the image $Img_i$. $G_i$ is the average gradient magnitude of the image $Img_i$, the calculation formula is shown in Eq. (2).

$$C_i = \frac{1}{N_i} \sum_{j=1}^{N_i} CS_j \tag{1}$$

$$\begin{cases} magnitude(x,y) = \sqrt{(grad_x(x,y))^2 + (grad_y(x,y))^2} \\ average\_gradient\_magnitude = \frac{1}{N} \sum_{i=0}^{N} magnitude(x_i, y_i) \end{cases} \tag{2}$$

We will further elaborate on how to utilize the metrics $C$, $N$ and $G$ to calculate the counting confidence $R$ in Section 3.2. Finally, the final counting result is calculated by taking the weighted sum of the counting results from all sequence images, as shown in Eq. (3). Each counting result $N_i$ is weighted by its softmax probability, which is derived from the counting confidence. $R_i$ represents the counting confidence of the image $Img_i$.

$$final\_count = \sum_{i=1}^{n} \left( \frac{e^{R_i}}{\sum_{j=1}^{n} e^{R_j}} \times N_i \right) \tag{3}$$

## 3.2 Modeling the aphid counting confidence evaluation system

Suppose we have $m$ sets of sequence images under interactive stirring actions. Each set contains $n$ time-ordered images, which can be represented as a two-dimensional matrix $IMG_{m \times n}$. As described in Section 3.1, we first calculate the factors influencing the counting confidence, including the detection confidence of aphid bounding boxes, the predicted number of aphids and the image clarity, resulting in corresponding two-dimensional matrices $C$, $N$ and $G$. At the same time, we compare the detection results with ground truth labels to calculate true positives (TP), false positives (FP), and false negatives (FN) for each aphid detection image, defining the counting confidence $R_{ij}$, as shown in Eq. (4).

$$R_{ij} = \frac{TP_{ij}}{TP_{ij} + FP_{ij} + FN_{ij}} \tag{4}$$

Next, we compute the average values of $C$, $N$, $G$ and $R$ for each time T across the $m$ sets of sequence images. This results in time-ordered arrays for $C^{'} = \left\{ C_1^{'}, C_2^{'}, ... C_n^{'} \right\}$, $N^{'} = \left\{ N_1^{'}, N_2^{'}, ... N_n^{'} \right\}$, $G^{'} = \left\{ G_1^{'}, G_2^{'}, ... G_n^{'} \right\}$, $R^{'} = \left\{ R_1^{'}, R_2^{'}, ... R_n^{'} \right\}$. After that, we conduct aphid counting confidence modeling. We use influencing factors (including $C$, $N$ and $G$) as independent variables and the confidence of the aphid counting results $R$ as the dependent variable to perform multiple linear regression analysis [11], aiming to determine the weights $w_C$, $w_G$ and $w_N$ of these influencing factors on the confidence of the aphid counting results, as shown in Eq. (5). It is worth noting that to ensure consistent data dimensions, we applied Min-Max Normalization to adjust all factor values to the same range [0,1] for each set of sequential data before conducting aphid counting confidence modeling.

$$R^{'} = w_0 + w_C \cdot C^{'} + w_G \cdot G^{'} + w_N \cdot N^{'} + \varepsilon \tag{5}$$

### 3.3 Aphid detection network

We propose a small object detection network based on Yolov5 for aphid detection, as shown in Fig. 2.

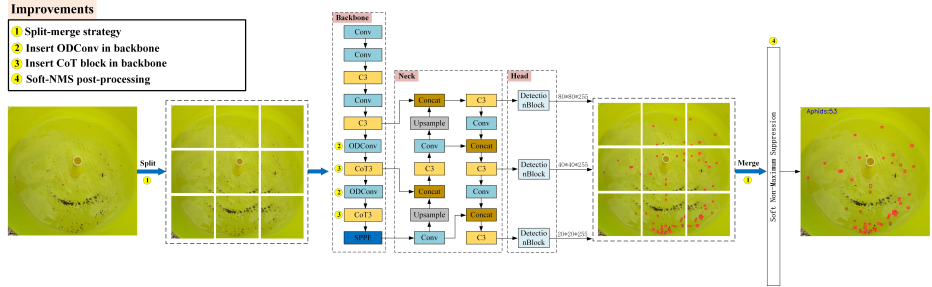

**Fig. 2.** The pipeline of our proposed aphid detection network.

Our proposed aphid detection network has four main improvements over the original Yolov5 (see Fig. 2): 1) Split-merge strategy. Inspired by Adam et al. [12], we divide the input image into 640x640 blocks with 20% overlap, feeding these blocks into the network one at a time. The detection results from these blocks are then merged to provide the final aphid counts. Unlike [12], we apply this slicing approach during the training phase as well, not just during inference, to further improve the model's performance in detecting small objects [13]. 2) Insert ODConv [14] in backbone. OD-

Conv introduces a multi-dimensional attention mechanism that dynamically adjusts the convolutional kernel across four dimensions. This flexibility allows the model to better capture relevant information in complex and dynamic scene. 3) Insert CoT3 block [15] in backbone. The CoT3 module combines the advantages of CNNs in local feature extraction with the strengths of Transformers in capturing global contextual relationships. It improves the model's overall learning capability for aphid detection. 4) Soft-NMS [16] post-processing. Soft-NMS improves upon traditional NMS by decaying the confidence scores of overlapping detection boxes rather than removing them, which helps retain potential true targets.

## 4 Experiments and results

To validate the effectiveness and superiority of our proposed aphid counting method, we first compared the performance of our proposed aphid detection network based on Yolov5 with that of the original Yolov5 for aphid detection. Then, we conducted comparative experiments on aphid counting using different methods, including static counting, maximum count under interactive stirring actions, our proposed counting method under stirring actions, and manual counting.

### 4.1 Dataset

**Dataset for aphid detection.** Six yellow water traps were placed in a sugar beet field to attract aphids, with images captured twice weekly from May to August 2024 using a 12MP smartphone camera. Images were taken at intervals from $T_0$ to $T_4$ (see Fig. 1), with a carpenter's tool used to stir the yellow water trap, and photos taken every two seconds. After each sequence of image collection, the yellow water traps were cleaned and refilled with water. Finally, a total of 570 images were collected, annotated using LabelImg, and split into training, validation, and test sets in a ratio of 8:1:1, resulting in 456, 57, and 57 images, respectively. The training and validation sets were then split into equal-sized blocks using a sliding window (640x640) and blocks containing aphids were filtered out for training the aphid detection model.

**Dataset for modeling and testing aphid counting confidence.** We collected and annotated 9 sets of sequence images under interactive stirring actions, with each set containing 9 images. Seven of these sets were used to build the aphid counting confidence model, while the remaining 2 sets were used for testing.

### 4.2 Implementation details

All experiments were conducted using Python 3.8.13 and PyTorch 1.12.1. During training, the aphid detection model used partially pre-trained Yolov5 weights to speed up convergence. The network input size was 640x640 with a batch size of 4, optimized using SGD with a 0.01 learning rate and 0.937 momentum, over 600 iterations. To prevent overfitting, the dataset was augmented with random translations, scaling, flipping, color adjustments, and mosaic augmentation.

### 4.3 Evaluation of aphid detection network

We performed aphid detection tests on the test set of the aphid dataset using both the original Yolov5 and our proposed aphid detection network. The comparison results are shown in Table 1.

**Table 1.** The comparison results of detecting aphids using different networks on the test set.

| Method | AP @0.5 (%) | AP @[0.5:0.95] (%) |
|---|---|---|
| Original Yolov5 | 40.9 | 17.2 |
| Ours | 74.8 | 44.1 |

As shown in Table 1, our proposed aphid detection network achieved 33.9% improvements in AP @0.5 and 26.9% improvements in AP @[0.5:0.95] compared to the original Yolov5. Thus, it significantly outperforms the original Yolov5.

### 4.4 Evaluation of aphid counting under interactive stirring actions

**Modeling and analysis results of aphid counting confidence.** We used seven sets of sequence images to build the aphid counting confidence model. The change over time T for each factor, including detection confidence of aphid bounding boxes $C$, predicted number of aphids $N$, image clarity $G$, and confidence of aphid counting $R$, is shown in Fig. 3. The results of multiple linear regression analysis are shown in Table 2.

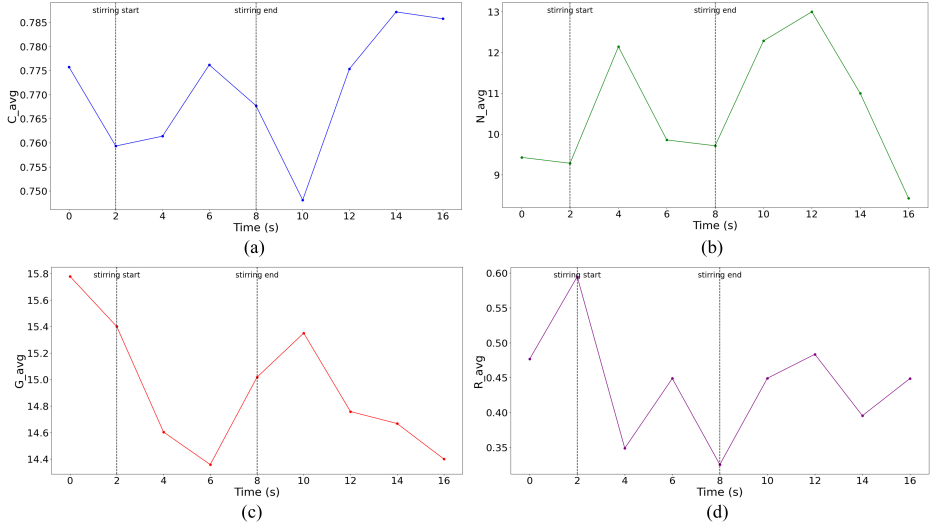

**Fig. 3.** The change in factors influencing aphid counting confidence model under stirring actions over time T.

Fig. 3 shows that the average detection confidence of aphid bounding boxes $C\_avg$ remains stable between 0.75 and 0.8. The average predicted number of aphids $N\_avg$ shows an overall trend of first increasing and then decreasing. This is because stirring brings some submerged aphids to the water surface. Image clarity $G\_avg$ first decreases, then increases, and finally decreases again due to the movement of water in the yellow trap caused by stirring. The average aphid counting confidence $R\_avg$ decreases during stirring due to image blurriness and increased false detections but increases once stirring ends.

**Table 2.** The results of multiple linear regression analysis.

| $w_0$ | $w_C$ | $w_N$ | $w_G$ | $\varepsilon$ |
|-------|-------|-------|-------|---------------|
| 0.3756 | -0.0023 | -0.1540 | 0.3205 | [-0.0988, 0.4189, -0.2181, 0.1326, -0.4799, -0.0104, 0.2767, -0.0963, 0.0752] |

From Table 2, we can see the calculated weights $w_C$, $w_N$, $w_G$ affecting aphid counting confidence, as well as the intercept term $w_0$ and the error term $\varepsilon$.

**Test results using the aphid counting confidence model.** Drawing on the calculations presented in Table 2, the aphid counting confidence for each image within the two test sequences is first computed. The final aphid count for each test sequence is then derived by applying a weighted sum of the aphid counts across all images, as formulated in Eq. (3). At the same time, we used static counting, maximum count under interactive stirring actions and manual counting methods to count the aphids in two sets of test sequence images. The comparison of aphid counting results is shown in Table 3.

**Table 3.** The comparison of aphid counting results using different methods.

| Group number | Static counting | Maximum count under interactive stirring actions | Ours | Manual counting |
|--------------|-----------------|--------------------------------------------------|------|-----------------|
| 1 | 2 | 17 | 9 | 10 |
| 2 | 15 | 23 | 18 | 21 |

From Table 3, our proposed counting method under interactive stirring actions significantly outperforms static counting, yielding counts that are 4.5 times higher in the first test set and three more in the second test set. However, it falls short of manual counting by one and three aphids, respectively, as not all pests in the yellow water trap are visible at the same moment during stirring. Additionally, the maximum count under interactive stirring actions tends to be overestimated.

## 5    Conclusion

This paper proposes a novel aphid counting method using interactive stirring actions and a counting confidence evaluation system. Stirring helps bring submerged aphids to the water surface, enabling more accurate counting. The counting confidence evaluation provides more reliable counting results. Experimental results show that our proposed counting method significantly outperforms static counting and closely matches manual counting. However, it is worth noting that this work is still in progress, particularly in terms of the dataset, which is currently insufficient with only 9 sets of sequence images. Future work will focus on collecting more data to improve the counting confidence evaluation system and further investigate the impact of stirring actions on counts, such as the duration and type of stirring.

## Acknowledgements

This work was supported by the Engineering and Physical Sciences Research Council [EP/S023917/1], the AgriFoRwArdS CDT, and the BBRO.